\pdfoutput=1

\documentclass[11pt]{article}

\usepackage[final]{acl}

\usepackage{times}
\usepackage{latexsym}

\usepackage[T1]{fontenc}

\usepackage[utf8]{inputenc}

\usepackage{microtype}
\usepackage{xcolor}
\usepackage{tikz}

\usepackage{dingbat}
\usepackage{amsmath}
\title{ED-FAITH: Evaluating Dialogue Summarization on Faithfulness}

\author{Sicong Huang\thanks{Work done during AI Residency at Meta} \quad Asli Celikyilmaz \quad Haoran Li \\
  Meta AI \\
  \texttt{sicong.huang0@gmail.com} \\
  \texttt{\{aslic, aimeeli\}@meta.com} \\
  }

\begin{document}
\maketitle
\begin{abstract}
Abstractive summarization models typically generate content 
unfaithful to
the input, thus highlighting the significance of evaluating the faithfulness of generated summaries. 
Most faithfulness metrics are only evaluated on news domain, \textit{can they be transferred to other summarization tasks?} In this work, we first present a systematic study of faithfulness metrics for dialogue summarization.
We evaluate common faithfulness metrics on dialogue datasets and observe that most metrics correlate poorly with human judgements despite 
performing well on news datasets. Given these findings, to improve existing metrics' performance on dialogue summarization, we first finetune on in-domain
dataset, then apply unlikelihood training on negative samples, and show that they can successfully improve metric performance on dialogue data. Inspired by the strong zero-shot performance of the T0 language model, we
further propose T0-Score -- a new metric for faithfulness evaluation, which shows consistent improvement against baseline metrics across multiple domains.
\end{abstract}




\section{Introduction}


Abstrative text summarization aims to condense a piece of text into a shorter version by distilling the information in the source text and 
rewriting it in a concise manner.
Recent advancements in pretrained language models \citep{NIPS2017_3f5ee243, NEURIPS2020_1457c0d6, JMLR:v21:20-074, lewis-etal-2020-bart, Zhang2020PEGASUSPW}
have enabled summarization systems to generate highly fluent and coherent summaries
on common summarization datasets such as CNN/DailyMail \citep{nallapati-etal-2016-abstractive}, XSum \citep{narayan-etal-2018-dont}.


Although news domain has long been the focus of summarization research, dialogue summarization, with many practical applications has gained more research attention lately. SAMSum \citep{gliwa-etal-2019-samsum} is the first large-scale dataset with human annotated abstractive summaries for chat-dialogue conversations. ConvoSumm \citep{fabbri-etal-2021-convosumm} created the first comprehensive benchmark for conversational summarization across diverse domains of news comments,
discussion forums, community question answering forums, and email threads.

Success in a summarization system's application is crucially dependent on its faithfulness: 
\textit{the factual alignment between the generated summary and the source}. 
\citet{kryscinski-etal-2020-evaluating} found that up to 30\% of generated summaries are affected by factual inconsistencies. \citet{tang-etal-2022-confit} studied types of factual errors generated by current models on popular dialogue summarization dataset and revealed hallucination issues. Thus having metrics that can reliably identify hallucinations and source-contradicting information becomes a critical step in summarization research.

Although commonly used to evalaute summarization models, metrics based on n-gram overlaps, such as ROUGE \citep{lin-2004-rouge}, BLEU \citep{papineni-etal-2002-bleu} and METEOR \citep{lavie-agarwal-2007-meteor}
are inadequate to measure a summary's faithfulness, because of their low correlation with human judgements \citep{NEURIPS2021_e4d2b6e6, deng-etal-2021-compression, maynez-etal-2020-faithfulness}. 
Even though many newly proposed metrics specifically target faithfulness
\citep{kryscinski-etal-2020-evaluating, wang-etal-2020-asking, NEURIPS2021_e4d2b6e6, deng-etal-2021-compression}
they are only evaluated on the news domain.
Because of the unique challenges inherent to dialogue data, i.e. spoken terms, special discourse structures, coreferences and ellipsis, 
etc. \citep{chen-etal-2021-dialogsum},
we suspect they will not be as effective in evaluating dialogue summaries right out-of-the-box.

In this paper, we evaluate the faithfulness aspect of common summarization metrics in the dialogue domain and investigate various approaches
to improve their reliabilities. Our contributions are: 
\vspace{-4mm}
\begin{enumerate}
    \item We analyze faithfulness metrics for dialogue summarization and find that metrics that perform well on one domain don't transfer directly to another domain.
    \vspace{-3mm}
    \item We improve existing faithfulness evaluation metrics for dialogue summarization by finetuning on in-domain data and applying unlikelihood training with negative samples.
    \vspace{-3mm}
    \item We propose a more general faithfulness metric, \textit{T0-Score}, and show that it 
    outperforms many metrics across multiple domains.
\end{enumerate}

\section{Related Work}
\label{sec:related}
\subsection{Summarization Metrics}
Common summarization metrics ROUGE, BLEU, and METEOR \citep{lin-2004-rouge, papineni-etal-2002-bleu, banerjee-lavie-2005-meteor}
are based on n-gram overlaps, which were shown to not correlate well with human judgements of factual consistency 
\citep{falke-etal-2019-ranking, kryscinski-etal-2020-evaluating}.
Newly proposed metrics aim to solve this problems with various approaches: 
FactCC \citep{kryscinski-etal-2020-evaluating} generates weakly supervised training data and treat consistency evaluation as an NLI task.
QAGS and FEQA \citep{wang-etal-2020-asking, durmus-etal-2020-feqa} model consistency evaluation as question generation and answering.
BERTScore \citep{bert-score} uses contextualized embeddings to compute similarities between the generated and the reference summaries.
CTC \citep{deng-etal-2021-compression} introduces information alignment that measure consistency at the token level.
BARTScore \citep{NEURIPS2021_e4d2b6e6} formulates evaluation of generated text as a text generation problem.

\subsection{Meta-Evaluation of Summarization}
\citet{maynez-etal-2020-faithfulness} conducted a human evaluation of hallucinated content in system generated summaries on 
XSum data and found textual entailment scores are best correlated with summary faithfulness.
\citet{fabbri-etal-2021-summeval} assembled SummEval, a collection of model-generated CNNDM summaries and their human judgement scores
along four axes including factual consistency. They found, on system-level, metrics using higher-order n-gram overlap such as ROUGE-3 are more 
effective. 
Similarly, \citet{pagnoni-etal-2021-understanding} composed FRANK, consists of generated summaries from CNNDM and XSum and corresponding
human annotations of factuality based on a typology of factual errors. They observe all tested metrics exhibit low correlations with human
judgements with the best metric FactCC achieving 0.3 Spearman correlation.
\citet{gabriel-etal-2021-go} proposed a meta-evaluation framework, GO Figure, which evaluates the sensitivity and validity of factual 
consistency metrics with only reference summaries. They found SummaQA, ROUGE-(2/3), and BERTScore perform better than ROUGE(1/L).

\section{Metrics and Data}
\label{sec:metrics}

\paragraph{Baseline Metrics} We compare 4 summarization metrics that are either widely used or shown to correlate well with human judgement in terms of faithfulness in news domain and seek further improvement of automatic metrics for faithfulness on dialogue summarization.


\textbf{ROUGE} \citep{lin-2004-rouge} measures summarization quality by counting n-gram overlaps between the hypothesis and the human written reference summary.

\textbf{BERTScore} \citep{bert-score} compares the similarity between the hypothesis and the reference by measuring the cosine similarities between each of
their token's contextualized embedding.

\textbf{CTC} \citep{deng-etal-2021-compression}, when evaluating \textit{consistency} with its discriminative model, works by predicting the probability of each token in the hypothesis
being consistent with the source. Training requires constructing negative training samples.


\textbf{BARTScore} \citep{NEURIPS2021_e4d2b6e6} when used to evaluate factuality, works by conditioning on the source and predicting the probability
of a hypothesis being generated with a seq2seq language model


\paragraph{Datasets}
We use SAMSum \citep{gliwa-etal-2019-samsum} dataset to study faithfulness metrics for dialogue summarization. The SAMSum corpus is a large-scale dialogue summarization dataset that contains 16k English daily conversations with corresponding summaries written by linguists. We use the human annotation of SAMSum summaries in ConFiT \citep{tang-etal-2022-confit} as our meta-evaluation dataset, where they generate summaries from six summarization models and collected faithfulness score on a scale of 1-10. We refer to this dataset as \textbf{MetaSAMSum}.




\section{Methods}
\label{sec:methods}
During evaluation, we found that off-the-shelf CTC and BARTScore perform poorly on MetaSAMSum (see "Vanilla" column of Table \ref{tab:in-domain}).
We hypothesize that the low performance of CTC and BARTScore on the SAMSum dataset is twofold. 
1. Neither of them are trained on dialogue data and their performance transfer poorly with shifted domain. 
2. BARTScore has never seen negative samples, so it struggles to recognize unfaithful summaries.
Thus, we design two approaches to improve them.

\subsection{In-domain Training}
\citet{NEURIPS2021_e4d2b6e6} have shown when evaluating the \textit{factuality} of generated text BARTScore finetuned on CNNDM performs the best. 
CTC is also trained on data constructed from CNNDM and XSum. However, since we find that they don't transfer well to evaluating dialogue summarization, based on hypothesis 1, we experiment with adapting CTC and BARTScore to the dialogue domain by training on in-domain data.


We use conversations from Pushshift Reddit \citep{Baumgartner_Zannettou_Keegan_Squire_Blackburn_2020} \footnote{an existing dataset extracted and obtained by a third party and made available on pushshift.io} as unsupervised in-domain training data.
Since Reddit conversations have no associated reference summaries, and results from CTC (in Table \ref{tab:overall}) have shown using extractive summaries on conversation data hurts performance, to train BARTScore, 
we generate fake references with T0 \citep{sanh2022multitask}, 
a multi-task seq2seq LM exhibiting strong zero-shot generalization abilities. For each conversation, we generate multiple
summaries with different prompts and pick the summary with the highest ROUGE-L score comparing against the source as the 
fake reference to prevent it from deviating too much from the source. 
In addition, We also investigate training directly on SAMSum, the task data.

\subsection{Unlikelihood Training with Negative Samples}

BARTScore works by predicting the probability of a hypothesis being generated given a source text. 
In the finetuning stage, standard MLE training aims to maximize the probability of positive tokens but does not explicitly minimize those of negative tokens. Based on hypothesis 2, i.e. without explicit guidance, it is difficult for the model to discern the unfaithful content,
we remedy this difficulty by leveraging unlikelihood loss \citep{Welleck2020Neural} to penalize negative tokens, 
teaching the model to assign low probabilities to unfaithful tokens.

Consider the sequence $S = (x_1, ..., x_T)$, $N$ is the set of indices of its negative tokens.
The loss for $S$ is given as
\begin{equation*}
    L_S = 
    \left\{\begin{array}{lr}
    \!\!-\sum\limits_{t}^{T} \log p_\theta (x_t|x_{<t}), & \!\!\text{if } S \text{ is pos}
    \\
    \!\!-\alpha \sum\limits_{t \in N} \log (1-p_\theta (x_t|x_{<t})), & \!\!\text{if } S \text{ is neg}
\end{array}\right.
\end{equation*}
Where $p_\theta$ is the model parameterized by $\theta$, $\alpha$ is a hyperparameter representing the weight of the unlikelihood loss.

Inspired by \citet{cao-wang-2021-cliff}, we generate three types of negative samples from SAMSum reference summaries.
\textbf{Swapent} shuffles entities of the same type, simulating wrong reference error which is a common error type for dialogue summarization.
\textbf{Maskent} randomly masks one entity of each type, and fill the masks with BART, simulating summaries with incorrect subjects or objects.
\textbf{Hallu} generates SAMSum summaries using BART trained on XSum with top \textit{p} sampling where \textit{p} = 1.0, simulating complete hallucinations.
We show examples of negative samples generated by our system in Appendix \ref{app:neg-sum}.

\section{Experiments \& Analysis}
\label{sec:exp&ana}
We report and discuss experiment results and new findings in this section. 
Each model is trained on 4 A100 GPUs using the AdamW optimizer with learning rate set to 2e-5. Trained with DDP, the effective batch size is 64.
All reported numbers are the average of 3 runs using different random seeds.

\subsection{In-domain Results}


\begin{table}
\centering
\begin{tabular}{|l|c|c|c|}
\cline{2-4}
\multicolumn{1}{c|}{}  & Vanilla    & w. Reddit & w. SAMSum \\ \hline
BScore                 & 0.1844 & 0.2667    & 0.3275    \\ \hline
CTC                    & 0.1245 & 0.0552    & 0.0692    \\ \hline
\end{tabular}
\caption{Spearman correlations of BARTScore (BScore) and CTC on SAMSum data with in-domain training. 
"Vanilla" uses \texttt{bart-large-cnn} and \texttt{D-XSum} for BARTScore and CTC respectively.}
\label{tab:in-domain}
\end{table}

Table \ref{tab:in-domain} shows training on Reddit conversations with fake references increases BARTScore's
correlation with human judgements by 45\%.
However, training on CTC's negative samples results in decreases in correlation. 
This suggests fine-tuning BARTScore on in-domain data even with weakly supervised summaries can boost its performance at evaluating
the faithfulness of dialogue summaries.
However, CTC did not generalize well to the new domain. 
We suspect this is due to CTC's negative generation pipeline being not suitable for dialogues, producing low-quality negative samples.

Nonetheless, training directly on task data i.e. SAMSum with human written references yields better
performance, despite SAMSum being 10 times smaller than unsupervised Reddit data.

\subsection{BARTScore with Negatives Ablation Results}

\begin{table}
\centering
\begin{tabular}{|c|c|c|c|r|}
\cline{2-4}
\multicolumn{1}{}{}        &  \multicolumn{3}{|c|}{Types of Negatives}     \\ \hline
ref         & swapent        & maskent          & hallu  &  correlations    \\ \hline
\checkmark  &                 &                 &             &  0.3275     \\ \hline
\checkmark  & \checkmark      &                 &             &  0.3414     \\ \hline
\checkmark  &                 & \checkmark      &             &  0.3400     \\ \hline
\checkmark  &                 &                 & \checkmark  &  0.3302     \\ \hline
\checkmark  & \checkmark      & \checkmark      &             &  0.3349     \\ \hline
\checkmark  & \checkmark      & \checkmark      & \checkmark  &  \textbf{0.3441} \\ \hline
\end{tabular}
\caption{Spearman correlations of BARTScore on SAMSum data. BART models are trained on SAMSum's human written references and different combinations of negative examples.}
\label{tab:unlikelihood}
\end{table}

We report ablation results of BARTScore with additional unlikelihood training on negative samples in Table \ref{tab:unlikelihood}, 
in which the weight of unlikelihood loss $\alpha$ = 0.1 and all models start from \texttt{bart-large-cnn}.
All models trained with unlikelihood loss on negative samples in addition to MLE show performance improvements.
When training with one type of negative, \textbf{swapent} contributes the most improvement among the three types of negative samples,
and \textbf{hallu} contributes the least.
Using all three types of negatives yields the best performance.

\subsection{T0-Score}
\begin{table}
\centering
\begin{tabular}{|l|c|c|c|}
\hline
Metrics      & SAMSum & CNN & XSum \\ \hline
CTC         & 0.1129 & 0.4293   & \textbf{0.3149}    \\ \hline
BARTScore   & 0.3441 & 0.3820   & 0.1705    \\ \hline
T0-Score (3B)    & 0.3049 & 0.4141   & 0.1829    \\ \hline
T0-Score (11B)      & \textbf{0.3780} & \textbf{0.4573}   & 0.1862    \\ \hline
\end{tabular}
\caption{Spearman correlations of CTC, BARTScore, and T0-Score, on SAMSum, CNNDM, and XSum.}
\label{tab:t0}
\end{table}

From previous ablations, we found that \texttt{bart-large-cnn} performs better than \texttt{bart-large} when finetuned on SAMSum for evaluation, which suggests that a more generalized model could do better as an evaluator. Because T0 is known for its strong zero-shot generalization abilities,
we experiment with using its generation probability of summary conditioned on source as the faithfulness metric, which we call T0-Score. Table \ref{tab:t0} shows results of 2 versions of T0-Score on meta evaluation datasets from multiple domains compared to other metrics. T0-Score shows performance gains even over the best
performing BARTScore on SAMSum. It also outperforms CTC on CNNDM, setting state-of-the-art performance. However, it falls short on XSum compared to CTC.

\subsection{Discussion}
\begin{table}
\centering
\begin{tabular}{|l|r|}
\hline
\multicolumn{2}{|l|}{\textit{Baseline Metrics}} \\\hline
ROUGE-1                                       & 0.2462                        \\ \hline
ROUGE-2                                       & 0.2626                        \\ \hline
ROUGE-3                                       & 0.2415                        \\ \hline
ROUGE-L                                       & 0.2461                        \\ \hline
BERTScore (roberta-large-mnli)                 & 0.2957                        \\ \hline
BERTScore (deberta-v2-xxlarge-mnli)           & 0.2974                         \\ \hline
CTC (D-XSUM)                                  & 0.1245                          \\ \hline
CTC (extractive on SAMSum)                       & 0.0253                        \\ \hline
BARTScore (bart-large)                        & 0.1315                        \\ \hline
BARTScore (bart-large-cnn)                     & 0.1844                        \\ \hline
\hline
\multicolumn{2}{|l|}{\textit{Our Metrics}} \\\hline
BARTScore (reddit)                              & 0.2667                        \\ \hline
BARTScore (unlikelihood)                         & 0.3441                        \\ \hline
T0Score (11B)                                    & 0.3780                        \\ \hline

\end{tabular}
\caption{Spearman correlation of all evaluated metrics}
\label{tab:overall}
\end{table}

Table \ref{tab:overall} summarizes results from all metrics together with our improvements. We see that most existing summarization faithfulness metrics perform poorly on dialogue data out-of-the-box (with BERTScore being an exception). BARTScore can be significantly improved when finetuned on in-domain or task data, and additional unlikelihood training with negative samples can further boost its performance. Since a metric's performance can be highly dependant on the domain that it is applied to, e.g. CTC is competitive on news but falls short on dialogues, we call for researchers to evaluate on multiple domains when proposing general automatic metrics for summarization evaluation.

\section{Conclusion}
\label{sec:conclusion}
We evaluate common faithfulness metrics on dialogue summarization and find that most of them exhibit low correlations
with human judgements. We experiment with ways to improve them by training on in-domain data with unlikelihood loss on negative samples,
and we show both of them can bring significant improvements. We call for more domain-aware use of evaluation metrics and more comprehensive evaluation over multiple domains when proposing new metrics.

\section{Limitations}

The negative samples we generate from handcrafted rules are limited to three types of common errors in auto-generated summaries.
To obtain more accurate faithfulness evaluations, more error types have to be proposed.
Or as an alternative, obtaining token-level human annotations is another expensive but viable option.

This study is conducted using SAMSum as a proxy to dialogue data. 
While our approaches are shown to perform well on SAMSum as a more casual chit-chat dialogue dataset, 
because of the lack of annotated data, we have not tested our approaches on datasets with formal and longer conversations. 
When more annotated data is available, evaluating on a more diverse set of dialogue datasets could make our arguments stronger.

\bibliography{anthology,custom}
\bibliographystyle{acl_natbib}

\appendix

\section{Generated Negative Samples} \label{app:neg-sum}
Table \ref{tab:ex-neg} shows examples of negative samples generated by our system.

\begin{table}[h!]
\begin{tabular}{l|p{5.5cm}}
\hline
Dialogue & Amanda: I baked  cookies. Do you want some? Jerry: Sure! Amanda: I'll bring you tomorrow :-) \\ \hline
Reference & Amanda baked cookies and will bring Jerry some tomorrow.                                    \\ \hline \hline
Swapent  & \colorbox{lightgray}{Jerry} baked cookies and will bring \colorbox{lightgray}{Amanda} some tomorrow.             \\ \hline
Maskent  & \colorbox{lightgray}{I have} baked cookies and will bring Jerry some tomorrow.                                     \\ \hline
Hallu    & \colorbox{lightgray}{Amanda: I baked cookies, and I} \colorbox{lightgray}{want to bring them to your house so} \colorbox{lightgray}{you can eat them!}         \\ \hline
\end{tabular}
\caption{An example SAMSum dialogue and its three types of negative summaries. Highlighted parts are negative tokens.}
\label{tab:ex-neg}
\end{table}

\end{document}